\documentclass{ecai} 

\usepackage{latexsym}
\usepackage{amssymb}
\usepackage{amsmath}
\usepackage{amsthm}
\usepackage{booktabs}
\usepackage{enumitem}
\usepackage{graphicx}
\usepackage{color}

\usepackage{dsbda}

\newcommand{\BibTeX}{B\kern-.05em{\sc i\kern-.025em b}\kern-.08em\TeX}

\begin{document}

\begin{frontmatter}

\title{Your Extreme Multi-label Classifier is Secretly a \\
Hierarchical Text Classifier for Free}

\author[A]{\fnms{Nerijus}~\snm{Bertalis}\orcid{0009-0000-4270-2308}}
\author[A]{\fnms{Paul}~\snm{Granse}\orcid{0009-0009-7813-5312}}
\author[A]{\fnms{Ferhat}~\snm{Gül}\orcid{1234-5678-9012-3456}} 
\author[A]{\fnms{Florian}~\snm{Haus}\orcid{0009-0008-5561-6773}} 
\author[A]{\fnms{Leon}~\snm{Menkel}\orcid{0009-0004-6058-7541}} 
\author[A]{\fnms{David}~\snm{Schüler}\orcid{0009-0000-7665-9707}} 
\author[A]{\fnms{Tom}~\snm{Speier}\orcid{0009-0006-2691-5921}} 
\author[B]{\fnms{Lukas}~\snm{Galke}\orcid{0000-0001-6124-1092}} 
\author[A]{\fnms{Ansgar}~\snm{Scherp}\orcid{0000-0002-2653-9245}
\thanks{Corresponding Author. Email: firstname.lastname@uni-ulm.de}}

\address[A]{Ulm University, Germany}
\address[B]{University of Southern Denmark, Odense, Denmark}

\begin{abstract}
Assigning a set of labels to a given text is a classification problem with many real-world applications, such as recommender systems. Two separate research streams address this issue. Hierarchical Text Classification (HTC) focuses on datasets with label pools of hundreds of entries, accompanied by a semantic label hierarchy. In contrast, eXtreme Multi-Label Text Classification (XML) considers very large sets of labels with up to millions of entries but without an explicit hierarchy. In XML methods, it is common to construct an artificial hierarchy in order to deal with the large label space before or during the training process. Here, we investigate how state-of-the-art HTC models perform when trained and tested on XML datasets and vice versa using three benchmark datasets from each of the two streams. Our results demonstrate that XML models, with their internally constructed hierarchy, are very effective HTC models. HTC models, on the other hand, are not equipped to handle the sheer label set size of XML datasets and achieve poor transfer results. We further argue that for a fair comparison in HTC and XML, more than one metric like F1 should be used but complemented with P@k and R-Precision. 
\textit{Source code is available on EasyChair.}
\end{abstract}

\end{frontmatter}

\section{Introduction}\label{sec:introduction}

Text classification is an essential task in text mining and natural language processing (NLP), which involves assigning pre-defined labels to input text, with applications in recommender systems, information retrieval,  sentiment analysis, and more~\cite{textclassificationsurvey,intro3,intro4,htc}. Since the introduction of text classification decades ago, many approaches have been developed, with a recent trend focusing on deep learning models~\cite{li2021survey}. 
When considering more than one possible output per example, a.k.a. multi-label classification, the literature is split into two streams. On the one hand, there is \textbf{hierarchical text classification (HTC)}, which aims to make use of a hierarchical structure among the classes \cite[inter alia]{hgclr}. On the other hand, there is \textbf{extreme multi-label text classification (XML)}, which aims to develop techniques to handle an extensive set of possible classes \cite[inter alia]{you2019attentionxml}. From a conceptual point of view, both families of approaches aim to tackle the same problem: multi-label text classification. \textbf{However, the literature is split in the sense that researchers employ different datasets and different evaluation metrics and do not consider the methods from the other side for comparison when developing new methods.} Interestingly, XML methods often use an artificial hierarchical structure to deal with the large label space. We question whether this harsh distinction of problems and methods can be alleviated and experiment with methods from XML applied to datasets used in HTC and vice-versa.

In \textbf{HTC}, the text classification models exploit the inherent information of a hierarchically structured set of classes~\cite{hiagm}. Furthermore, HTC utilizes a top-to-bottom approach for path construction and hierarchical relationships to improve classification performance~\cite{htc}. A common approach to the taxonomic hierarchy in HTC is to model a graph or tree, where each node represents a label to be classified~\cite{hgclr}. In the field of HTC, there have been many different approaches that incorporate hierarchical information. For instance, independently processing text and structure of label hierarchy and then merging the two elements into a joint representation~\cite{himatch}. By doing so, it is aimed to reach a combined representation of text and hierarchy that is beneficial for classification~\cite{hiagm}. Another usage of hierarchy for text classification within HTC can be narrowed down to local and global approaches~\cite{hiagm}, where local approaches aim to create classifiers at each level of the hierarchy by using strategies such as CNNs. On the other hand, global strategies approach HTC as a form of multi-label classification by integrating hierarchical information directly into models to rely on the inherent structure and relationships within the hierarchy to improve classification accuracy~\cite{htc}. 

In \textbf{XML}, the main objective is to be able to handle large datasets with a large number of labels. XML is an advanced text classification domain that focuses on assigning a relevant subset of labels to an instance, where the number of labels poses serious computational and statistical challenges, making it difficult to create an XML method that is efficient to train and accurate~\cite{xr-transformer,you2019attentionxml}. XML methods can be broadly categorized into sparse linear models, embedding-based methods, and transformer-based models~\cite{xr-transformer}. Initially, sparse linear models laid the foundation. They use linear algorithms and efficient partitioning to work with extensive label spaces by implementing hierarchical trees and multi-label classification techniques such as one-versus-all~\cite{XMCsparse}. 
Subsequently, embedding-based techniques were introduced to work with neural networks and aimed to learn semantic representations, often combined with sampling for scalability~\cite{you2019attentionxml}. 
Recent developments make use of pre-trained transformer language models to capture complex semantic relationships within the data~\cite{xr-transformer}. 

So far, it is unclear how the XML and HTC techniques would perform in unconventional settings, namely when XML methods encounter hierarchical data and HTC methods work with vast amounts of labels. Driven by a cross-domain evaluation, this study aims to identify opportunities for improvement within HTC and XML approaches and potential limitations.
We expect to gain key insights on the performance and usability of HTC and XML approaches and potential new applications in the field of multi-label text classification.

Until now, HTC and XML have existed as two distinct streams of research in multi-label text classification. 
Both streams aim to train a classifier on a given dataset containing texts and their respective class labels to predict relevant labels for an input. What distinguishes these two approaches is, firstly, that HTC models are additionally provided with a fixed label taxonomy. 
XML methods create such a hierarchy through clustering at runtime. Secondly, the number of labels in the HTC datasets is much smaller than in the XML datasets. 
A few hundred in the case of HTC, but up to millions when it comes to XML. 
For that reason, there has been no research or knowledge about model performance when applied to datasets from the other domain. 
To gain a deeper understanding of the individual capabilities of state-of-the-art models from both sides, this separation has to be dissolved by making the datasets from one side accessible to models from the other side without structural information loss or unfeasible computational costs. 
A further distinction is that the evaluation measures differ between the HTC domain and the XML domain. HTC commonly employs Micro-/Macro-F1, while XML uses ranking-based metrics such as P@k.  

\paragraph{Contribution}
So far, it was unclear whether the XML methods would be applicable to HTC datasets. We contribute a procedure on how to overcome the barriers of different dataset types and metrics, allowing future researchers and practitioners to employ XML methods for HTC datasets and vice versa. To facilitate such a comparison, we have harmonized the metrics and suggest using R-precision.
We show how the models can be applied to the datasets of the respective other research stream.
Our results show that XML models, specifically CascadeXML, achieve comparable or better results than HTC models when tested on datasets from both domains for the performance measures P@k and R-Precision. This allows us to argue that XML models are more versatile.
Our results suggest that XML and HTC are closer to each other than one would expect on the basis of a clearly divided literature. 
Our main finding is that XML models -- despite not making use of the manually curated hierarchies of HTC datasets -- are competitive to state-of-the-art HTC methods.

\arxivonly{
\paragraph{Organisation}
In Section~\ref{sec:relatedwork}, we summarize related work, subdivided into HTC and XML. In Section~\ref{sec:methods} we describe how we bring the two worlds together. Here we focus on the modification of benchmark datasets from one world in order to use them in the other. The experimental apparatus is described in Section~\ref{sec:experimentalapparatus}, where preprocessing steps, the general procedure, hyperparameter optimization, and performance measures are explained. An overview of the obtained results is given in Section~\ref{sec:results}, which are then discussed in Section~\ref{sec:discussion}.
}

\section{Related Work}
\label{sec:relatedwork}
Generally, the state of the art in single and multi-label text classification is fine-tuning a pre-trained encoder-only language model~\cite{li2021survey,galke2023really} such as BERT~\cite{devlin2019bert}.
These models even outperform generative large language models such as GPT-4~\cite{DBLP:journals/corr/abs-2303-08774-gpt4} and Llama-2~\cite{DBLP:journals/corr/abs-2307-09288-llama2} in accuracy while offering advantages of smaller model size and less training time~\cite{sun2023text, yu2023open}. They also improve on recently proposed graph-based methods~\cite{galke2023really}. 
Therefore, we focus on models based on pre-trained transformers as well as common baseline methods. 
We first discuss the state-of-the-art in HTC followed by XML.
For further reading, we refer to surveys like~\citeauthor{htc_review}~\citeyearpar{htc_review}.

\paragraph{Hierarchical Text Classification (HTC)}

Given a hierarchy of class labels organized as a tree (in short, a label tree) and an input text, the task of HTC is to produce a set of labels from the label tree that accurately represents the input text. Approaches in HTC are often distinguished as local and global~\cite{hiagm}, where local approaches focus on building appropriate classifiers for each node or each level in the label tree, whilst global approaches employ a single classifier for the whole hierarchy. 
The state-of-the-art models HGCLR and HBGL follow the global approach.

HGCLR~\cite{hgclr} is a model that effectively integrates the label hierarchy with the semantics from the text and labels using a contrastive learning framework. It employs BERT~\cite{devlin2019bert} as the text encoder and a modified Graphormer~\cite{graphormer} to capture hierarchical structures.
The key distinction in HGCLR's modified Graphormer is the combination of a label embedding and a name embedding.
The former is an embedding of the vertex in the hierarchy, and the latter is a textual embedding of the label string. 

HBGL~\cite{hbgl} removes the auxiliary graph encoder and uses BERT~\cite{devlin2019bert} for both text and graph encoding. The model consists of two separate training stages using the same BERT model. 
In the first stage, the approach learns label embeddings based on random walks through the hierarchy, while keeping the BERT model frozen.
Learning is performed using a modified version of BERT's masked language modeling objective and a custom attention mask to represent the hierarchy.
In the second stage, and based on these initial label vectors, the BERT model is trained to sequentially generate class labels based on the text input
with an objective similar to sequence-to-sequence fine-tuning~\cite{s2s-ft}.

Other models that leverage hierarchical information that work similarly to HGCLR are HPT~\cite{hpt}, which employs different graph encoders such as graph attention~\cite{gat}, and HALB~\cite{DBLP:journals/kbs/ZhangLSXTH24-halb}, which replaces the classification loss with an asymmetric loss and adds another loss for separating samples with similar representations but different labels.
HILL~\cite{hill2024} applies a hierarchy-aware contrastive loss, too, using BERT as a text encoder and a separate graph encoder.
Earlier methods did not rely on transformer models, such as HiAGM~\cite{hiagm}. 
However, leveraging pre-trained transformers has generally shown to enhance performance~\cite{galke2023really,htc-info-max,himatch} and resembles the state of the art in HTC with HBGL~\cite{hbgl}.

\paragraph{eXtreme Multi-label Classification (XML)}
XML, like HTC, aims to assign a subset of labels to a given text and leverages a hierarchy to improve accuracy and reduce training complexity. However, there are two main differences: First, the label spaces usually are much larger than in HTC. Second, the hierarchies do not come from the data as in HTC but are artificially constructed in XML.

There are two main aspects along which XML techniques can be categorized: First, there are conventional machine learning methods that employ sparse features such as bag-of-words (BoW) features from text. Second, there are deep learning methods that operate on the raw text data. XML algorithms in the first category can be divided into the following three subcategories: one-vs-all~\cite{babbar2016dismec, ppdsparse}, label-tree-based~\cite{parabe2018}, and embedding-based~\cite{sparselocalembeddings2015} methods.
Recently, deep learning methods from the second category that operate on the full sequence of text inputs (rather than a bag of words) have become increasingly popular~\cite{kharbanda2022cascadexml,xr-transformer}. They fine-tune pre-trained transformer models like BERT, RoBERTa~\cite{liu2019roberta}, or XLNet~\cite{yang2020xlnet} and have shown promising results in both prediction accuracy and model size. 
Strong performers in XML classification, among others, are AttentionXML~\cite{you2019attentionxml} using BiLSTM~\cite{DBLP:journals/tsp/SchusterP97} and InceptionXML~\cite{kharbanda2022inceptionxml} for extreme short text classification. 
LightXML~\cite{lightxml} employs an ensemble of BERT, RoBERTa, and XLNet together with dynamic negative label sampling via generative cooperative networks.

We discuss in detail the two strongest models for XML, which we consider for our study.
The XR-Transformer~\cite{xr-transformer} is a recursive architecture where the transformer layers are connected with a gradually refined increase in label resolutions. 
To accomplish this, a hierarchical label tree is formed by repeatedly applying a $k$-means clustering algorithm on the label features. The feature of a label $l \in L$ is computed by averaging the TF-IDF 
representations of the documents that carry a label $l$. 
Subsequently, the model is trained on each level of the tree in a top-down manner. 
At each level, a three-step algorithm is executed. 
The first step involves selecting the top-$k$ clusters from the parent level. 
These clusters are then used in the second step, where the model is fine-tuned on the label resolution of the current level. 
Finally, the top-$k$ clusters from the current level are ranked. 
XR-Transformer uses the same ensemble as LightXML but is 17x faster with similar accuracy.

CascadeXML~\cite{kharbanda2022cascadexml} addresses the complexity of XR-Transformer and outperforms the aforementioned approaches on the P@k metric. 
It uses an ensemble similar to XR-Transformer and LightXML.
CascadeXML takes advantage of the multi-layer architecture of transformers and does not treat them as black-box. 
Instead, CascadeXML trains a single transformer across multiple resolutions, allowing it to use attention maps specific to the current tree resolution. 
The method clusters all given labels and creates an HLT as in XR-Transformer. 
A shortlisting process is performed to select the highest-scoring meta-labels, which are referred to as level-one meta-labels. 
This process is continued until the final level is reached, which, in contrast to XR-Transformer, carries out classification at full label resolution.

\section{Transfer XML $\leftrightarrow$ HTC}\label{sec:methods}
Due to the different nature of tasks tackled, HTC and XML studies use different benchmark datasets. 
On the HTC side, widely used benchmark datasets are Web of Science~\cite{ds-wos}, The New York Times Annotated Corpus~\cite{ds-nyt}, and RCV1-v2~\cite{ds-rcv1v2}. On the other hand, Wiki10-31K~\cite{ds-wiki1031}, AmazonCat-13K~\cite{ds-amazon}, and Amazon-670K~\cite{ds-amazon} are commonly used in the XML literature. Note that these datasets differ in size, number of classes, and evaluation strategy for HTC and XML models. HTC models use datasets with smaller label sets and fewer data points than XML datasets and evaluate performance by calculating the classification metrics Micro-F1 and Macro-F1 scores. In contrast, XML models are commonly evaluated with ranking-based metrics, such as P@k and PSP@k~\cite{PSPatK} metrics. This poses challenges regarding the comparability of the two streams of research. 

In more detail, Micro-F1 and Macro-F1 scores are calculated based on hard yes or no classification decisions. P@k and PSP@k are ranking-based metrics that have an arbitrary cut-off point (as specified by $k$) after the highest-ranking $k$ predictions. This hinders computing an F1 score over the full range of classes. 
For a fair comparison, we decided to employ the P@k metric for the HTC models and additionally consider the R-Precision metric~\cite{eval-in-ir} for models from both domains to ensure that the results can be fairly compared. We describe this in more detail in Section~\ref{sec:measures}.

\subsection{Bidirectional Transfer of Datasets}
To answer our main research question of a bidirectional transfer of methods between HTC and XML, we need to convert the datasets to make the respective other methods applicable. In the following, we describe how the HTC hierarchies are collapsed in order to apply XML methods to these datasets. The other direction of constructing a label hierarchy for HTC models to work with XML datasets is described in Section~\ref{sec:methods:xml_to_htc}, including adjustment of model parameters and handling of large datasets exceeding the limits of the models.

The XML methods do not require a pre-defined label hierarchy so that the pre-defined hierarchy of the HTC  can simply be discarded.  After this step, however, there are two ways to complete the transfer. The first option is to transfer all the labels in the label set, and the second option is to transfer only the labels that appear as leaf nodes in the original hierarchy. These are the most specific labels. Note that depending on the hierarchical dataset, these sets are not necessarily disjoint. Our main experiments focus on the first version. Since HTC models train and predict with internal nodes as well as leaf nodes, we find that this version is most suitable to facilitate a fair comparison (no changes to the ground truth of assigned labels). 
In addition, CascadeXML and XR-Transformer need to construct a hierarchical label tree (see Section~\ref{sec:relatedwork}) for these datasets in a preprocessing step. How exactly this is done depends on the specific models. The construction of such hierarchical label trees is described in detail in Section~\ref{sec:methods:xml_to_htc}.

\subsection{Harmonizing Performance Metrics}
\label{sec:measures}
The performance of the models is measured differently in the domains of HTC and XML. 
HTC models typically use the classification metrics F1-Micro and F1-Macro~\cite{f-score}.
F1-Micro and F1-Macro are different ways of aggregating the F1 score~\cite{f-score}.
F1-Macro gives equal weight to each class regardless of frequency, and F1-Micro gives more weight to frequent classes. 
F1-Marco is calculated by averaging the F1 score over every class, and F1-Micro is calculated by treating every instance equally and computing the harmonic mean of the global (micro) precision and recall across all classes.

In contrast, XML models use the ranking-based metrics Precision@k (P@k) and Propensity Scored Precision@k (PSP@k)~\cite{PSPatK}.
P@k measures how well a model accurately identifies relevant elements within its top-$k$ predictions. It quantifies how many top-$k$ predictions are also marked as relevant in the ground truth. P@k is formally defined as 
$\operatorname{P@k} = \frac{1}{k} \sum_{i=1}^{k} \text{Relevance}(i) $
where Relevance(i) is an indicator function that is $1$ if the $i$-th item is relevant and $0$ otherwise.
PSP@k does the same, but has a special emphasis on tail labels and rewards their accurate prediction, as they are often challenging to predict~\cite{PSPatK}. Tail labels refer to those labels that occur infrequently across the dataset~\cite{longtaildistribution}. This metric is particularly relevant in research on XML because the label sets are very large, which means that tail labels occur often.

For a meaningful comparison of the performance of models across the HTC and XML domains, it is necessary to define a common evaluation metric.
An option is the R-Precision metric, which is an adaptation of the P@k metric that does not set a fixed $k$ but rather sets $k$ to the number of relevant labels in the ground truth for every data point that is being evaluated~\cite{eval-in-ir}. 
It is defined as
$\operatorname{R\text{-}Precision} = \frac{1}{r} \sum_{i=1}^{r} \text{Relevance}(i)$
where $r$ is the number of relevant labels in the ground truth for the data point that is evaluated. 
The metric is appealing as one does not need to consider multiple values of $k$ as in $P@k$.
Calculating these ranking-based metrics for the classification-based HTC models is usually possible because the methods learn probabilities for each label. 
This means that they internally also employ a ranking (e.g., it can be extracted from the logits), which can be utilized to calculate P@k scores and the R-Precision.

Another possible way of establishing a common metric would have been to modify the ranking-based XML models to make hard classification decisions (whether one specific class will be assigned or not), which would allow the calculation of classification metrics such as the F1 score. 
To do this, one would need to take the relevance ranking produced by the XML models and determine the cut-off point above which labels are classified as relevant. 
However, setting this cut-off point is not trivial and would need to be optimized separately for each dataset. In fact, this conversion of ranking to hard decisions is a research endeavor of its own, e.g., see \citet{tang2009large}, but usually not considered in XML papers. 
It should also be noted that it is not feasible to use a dynamic cut-off point based on the ground truth because this would not be available when the model is deployed. Even if one would take the approach, for example, of using the R-Precision metric and setting the cut-off point to the number of relevant labels in the ground truth for each query, that would entail that precision and recall would be mathematically equivalent, which would make the calculation of F1 scores obsolete~\cite{eval-in-ir}. 

For these reasons, we decided to consistently employ the ranking-based metrics $\operatorname{P@k}$ and $\operatorname{R-Precision}$ to facilitate our comparison. We excluded the PSP@k metric, due to the considerably smaller size of the label sets in HTC in comparison to the label sets in XML. This makes the problem of tail labels less relevant for the HTC methods.

\subsection{Transfer of XML Datasets for HTC Models}
\label{sec:methods:xml_to_htc}

In order for HTC methods to work with XML datasets, a hierarchical label tree as constructed by the XML methods is needed to serve as the hierarchy for the HTC methods.
In order to generate the hierarchical label tree, we employ a clustering method similar to CascadeXML. 
Each label in the set is assigned a numerical value based on the feature aggregation of its positive instances (PIFA)~\cite{kharbanda2022cascadexml}, which is computed as follows. 
For a label $l_i$ and the set $D_i$ of documents labeled with $l_i$ in the ground truth, PIFA is defined as $\operatorname{PIFA}(l_i) = \sum_{d\in D_i} \frac{\text{TF-IDF}(d)}{\operatorname{length}(d)}$.
Subsequently, a balanced $k$-means algorithm is applied recursively on these numeric label representations a total of $n$ times until a hierarchical label tree of depth $n$ is created. 
The difference between our method to transfer XML datasets to HTC models and the application of CascadeXML to the XML datasets is that we normalize the TF-IDF vectors by dividing them by the number of words in the documents, while CascadeXML takes the norm of the resulting embedding. This was done to reduce the influence of text length on clustering.
This means that labels are clustered together not only when they are relevant to documents with similar features but also when the labels have similar relevance frequencies. 

Note that by generating a hierarchy in this manner, we introduce new labels to the dataset. To allow a meaningful comparison, we use these new intermediate labels in training but remove them during classification. The true labels from the original dataset are now all leaf nodes within the generated hierarchy, even if a label was previously a parent node higher up in the original hierarchy. The branching factors and depths of these trees that were used in our experiments were selected on the basis of CascadeXML. 
The depth branching factors are summarized in Table~\ref{tab:branching_factors}.  

\begin{table}[h]
\centering
\caption{Branching factors for each level L1 to L3 (and L4, respectively) of the hierarchical label tree in XML datasets.}
\begin{tabular}{c|cccc}
\toprule
\textbf{Dataset} & \textbf{L1}  & \textbf{L2}  & \textbf{L3}  & \textbf{L4} \\
\midrule
Wiki10-31K  & 512 & 8 & 7.6 &   \\
AmazonCat-13K & 256 & 8 & 6.3 &\\
Amazon-670K & 1024 & 8 & 8 & 10.2\\
\bottomrule
\end{tabular}
\label{tab:branching_factors}
\end{table}

Both HBGL and HGCLR use this hierarchy as input to their underlying encoder within the model. In the case of HBGL, this encoder is BERT, while in HGCLR it is Graphormer. Each label (node) in the hierarchical label tree is reduced to a node embedding and serves as an input token for the encoder. The tree structure is preserved by using a custom attention matrix, whereby nodes can only attend to successive nodes.

The encoders have an inherent limit regarding the maximum number of input tokens. 
Since each label acts as an input token, all of the XML datasets surpass this limit. 
Furthermore, the quadratic complexity of the attention mechanism worsens the issue with large input sizes. To address this limitation, we segment the hierarchical label tree into smaller sub-trees, each maintaining the properties of a hierarchical label tree.

We perform this segmentation using depth-first search (DFS), exclusively traversing downward to gather just the leaf nodes and their corresponding preceding nodes. This method ensures that each leaf node in the original tree remains a leaf node in the new sub-tree. Additionally, the number of labels in each new sub-tree is kept within the token limit of BERT of 512 tokens.
By doing so, we might lose context between certain nodes, as they will not be aware of each other within the attention mechanism of the encoder. 
However, large parts of the general hierarchical structure should still be preserved, as distant nodes generally interact less with each other, \eg an article about sports is less likely to contain a lot of keywords related to astrophysics.

These encoders provide a refined embedding for each label. 
Although BERT has a maximum input token limit of 512, which might imply a restriction on the number of labels that can be input at once, this is not an issue. 
For HGCLR, label embeddings are not directly fed into BERT. 
For HBGL, the label embeddings are aggregated within each hierarchy level, and since the number of hierarchy levels is typically much lower, the classification process is not restricted by the number of labels.
However, compressing each label from a hierarchy level within HBGL into a single token will likely diminish the performance of the model.

To apply the HTC models on the XML datasets AmazonCat-13K and Amazon-679k, the number of training examples is set to $30k$ and another $5k$ for validation, while the test set is left unchanged. 
This is needed to allow for a reasonable training time.

\section{Experimental Apparatus}
\label{sec:experimentalapparatus}
\paragraph{Datasets}
\label{sec:datasets}
We consider six benchmark datasets, three HTC and three XML, see Table~\ref{tab:datasets}. 
The HTC datasets are Web of Science (WoS)~\cite{ds-wos}, The New York Times Annotated Corpus (NYT)~\cite{ds-nyt}, and RCV1-v2~\cite{ds-rcv1v2}.
Commonly used XML datasets are  Wiki10-31K~\cite{ds-wiki1031}, as well as AmazonCat-13K and Amazon-670K~\cite{ds-amazon}. 
All datasets are available publicly or after signing a license; NYT, while no longer licensable, is used under an existing valid license. We use NYT because it is frequently applied in HTC research.  

\begin{table}
\caption{Summary statistics of the HTC and XML datasets. Amount of labels and data points within each dataset. Avg($L_i$) is the average number of classes.}
\resizebox{0.5\textwidth}{!}{
    \begin{tabular}{l|rrrrr}
        \toprule
        \textbf{Dataset} & \textbf{\#Labels} & \textbf{\#Train} & \textbf{\#Val} & \textbf{\#Test} & \textbf{Avg($L_i$)}\\
        \midrule
        \multicolumn{6}{c}{HTC} \\
        \midrule
        WoS & 141  & 30,070 & 7,518 & 9,397 & 2,00 \\
        NYT & 166 & 23,345 & 5,834 & 7,292 & 7,60\\
        RCV1-v2 & 103 & 20,833 & 2,316 & 781,265 & 3,24\\
        \midrule
        \multicolumn{6}{c}{XML} \\
        \midrule
        Wiki10-31K & 30,938 & 14,146 & - & 6,616 & 18,64\\
        AmazonCat-13K & 13,330 & 1,186,239 & - & 306,782 & 5,04\\
        Amazon-670K & 670,091 & 490,449 & - & 153,025 & 5,45\\
        \bottomrule
    \end{tabular}
    }
    \label{tab:datasets}
\end{table}

\paragraph{Preprocessing}\label{sec:preprocessing}
For preprocessing, we follow the procedures described within the specific models HGCLR, HBGL, XR-Transformer, and CascadeXML. When preparing HTC datasets for the XML models and vice versa, we refer back to the dataset transfer described in Section~\ref{sec:methods}.

\paragraph{Procedure}\label{sec:procedure}

Our first step is the reproduction of published results for all models we use. 
Second, we convert the HTC datasets (WoS, NYT, RCV1-V2) by removing the hierarchy and vice versa, injecting a hierarchy into the XML datasets (Wiki10-31K, AmazonCat-13K, Amazon-670K) via clustering. 
With all preparations done, all four models are run on the respective datasets adjusted to their domain. Results are averaged over five runs.

\paragraph{Hyperparameter Optimization}
\label{sec:hyperparameteroptimisation}
For the reproduction of the results for each model, we used the hyperparameters provided by the authors, as those have already been tuned. 
When transferring methods to the datasets of the respective other domains, we chose the hyperparameters of the most similar datasets. 
For the XML methods on HTC datasets, we use those hyperparameters that each respective paper reports for the smallest XML dataset: Wiki10-31K.

\section{Results}\label{sec:results}\label{sec:results-rq2}

We report the performance of HTC and XML models on datasets from the HTC stream of research.
The results for the other direction, \ie applying HTC and XML models to XML datasets can be found in Section~\ref{sec:results-rq1}.

\subsection{HTC and XML methods on HTC datasets}\label{xml-on-htc-datasets}

On the NYT dataset (see Table~\ref{tab:nyt}, top) CascadeXML performs best on all metrics except P@5 with an R-precision of $84.61$, outperforming both HTC methods and XR-Transformer. 
It is followed by HGCLR with a difference in R-precision of $0.39$. 
HGCLR also performs best for P@5 with $71.74$ and a difference of $0.26$ to CascadeXML. 
For R-Precision, CascadeXML wins $8.68$ points over HBGL and $2.06$ points over XR-Transformer. 
XR-Transformer also beats HBGL in every metric, improving on HBGL by $6.62$ points in R-Precision. 
HBGL** is slightly stronger with an XML-style taxonomy than with the standard taxonomy of the dataset. 
Compared to its standard version, it gains $1.79$ points in R-precision and $2.05$ points in P@1.

\begin{table}[t]
\small
  \centering
  \caption{Results for the HTC datasets, namely New York Times Annotated Corpus (top), Web of Science (middle), Reuters Corpus Volume 1 (bottom).
  HBGL** indicates that the semantic hierarchy, as provided by the datasets in HTC, has been replaced by a synthetic hierarchical label tree as it is computed in XML.
  }
    \begin{tabular}{l|ccccc}
        \toprule
        \multicolumn{6}{c}{NY-Times} \\
        \midrule
        Method & P@1 & P@2 & P@3 & P@5 & R-Prec\\
        \midrule
        HGCLR & 95.02 & 93.16 & 83.98 & \textbf{71.74} & 84.22  \\
        HBGL & 83.81 & 83.26 & 75.37 & 64.87 & 75.93 \\
        HBGL** & 85.86 & 84.86 & 77.08 & 65.63 & 77.72 \\
        \midrule
        XR-Transf. & 92.56 & 91.50 & 82.73 & 70.85 & 82.55 \\
        CascadeXML & \textbf{95.93} & \textbf{93.56} & \textbf{83.94} & 71.48 & \textbf{84.61}\\
    \end{tabular}
\label{tab:nyt}

  \begin{tabular}{l|ccccc}
    \toprule
    \multicolumn{6}{c}{Web of Science} \\
    \midrule
    Method & P@1 & P@2 & P@3 & P@5 & R-Prec\\
    \midrule
    HGCLR & \textbf{91.05}  & \textbf{86.44} & \textbf{60.27} & \textbf{37.48} & \textbf{86.44} \\
    HBGL & 85.97 & 86.02 & 57.62 & 34.57 & 86.02 \\
    HBGL** & 87.21 & 86.12 & 57.49 & 34.49 & 86.12 \\
    \midrule
    XR-Transf. & 81.39 & 72.98 & 54.10 & 35.00 & 72.98 \\
    CascadeXML & 88.99 & 83.38 & 58.44 & 36.40 & 83.38\\   
  \end{tabular}
  \label{tab:wos}

    \begin{tabular}{l|ccccc}
    \toprule
    \multicolumn{6}{c}{RCVI-V2} \\
    \midrule
    Method & P@1 & P@2 & P@3 & P@5 & R-Prec\\
    \midrule
    HGCLR & 96.78 & \textbf{93.05} & \textbf{83.40} & \textbf{57.86} & \textbf{90.49}  \\
    HBGL & 91.51 & 88.96 & 80.43 & 54.57 & 85.19 \\
    HBGL** & 93.92 & 90.20 & 80.56 & 54.08 & 85.86 \\
    \midrule
    XR-Transf. & 97.09 & 92.39 & 82.36 & 56.52 & 88.54 \\
    CascadeXML & \textbf{97.31} & 92.54 & 82.61 & 57.44 & 89.01\\
    \bottomrule
  \end{tabular}
\label{tab:rcv1}
\end{table}  

The results on the WoS dataset (see Table~\ref{tab:wos}, middle) show a different picture. HGCLR performs best across all metrics, outperforming HBGL by $0.42$ points, CascadeXML by $3.06$ points, and XR-Transformer by $13.46$ points in R-Precision, respectively. 
In contrast to the WoS and RCV1 datasets, HBGL has the second-highest R-precision, beating CascadeXML by $2.64$ points. 
XR-Transformer is well behind all other methods. 
Note that for WoS, each data point has exactly two labels (see Table~\ref{tab:datasets}), so the P@2 equals the R-precision . 

The results for the RCV1-V2 dataset (see Table~\ref{tab:rcv1}, bottom) are similar to those for the NYT dataset. 
HGCLR performs best on P@2, P@3, P@5, and R-precision with a small margin over CascadeXML. 
Regarding P@1, CascadeXML has the upper hand with a difference of $0.53$ points to HGCLR. 
For the other metrics, the improvements of HGCLR over CascadeXML vary from $0.51$ (P@2) to $1.48$ (R-Precision). 
The differences between CascadeXML and XR-Transformer are small for all metrics ($< 1.0$ point) and HBGL is outperformed by both methods. Again, the results of HGCLR and CascadeXML are close, followed by XR-Transformer and HBGL.

\subsection{HTC and XML methods on XML datasets}
\label{sec:results-rq1}
\label{sec:htc-on-xml-datasets}

On the Wiki10-31K dataset (results are shown in Table~\ref{tab:wiki10-31k}, top) XR-Transformer performs best across all metrics with an R-Precision of $38.93$ and is followed by CascadeXML with a difference of $3.17$ points. 
XR-Transformer beats HBGL by far with differences up to $35.02$ points (P@1) and $12.10$ points in R-Precision.

\begin{table}[t]
\small
  \caption{Results for the XML datasets, namely Wiki10-31K (top), AmazonCat-13K (middle), and Amazon-670K (bottom). 
  For methods marked with \textdagger{} no results could be obtained due to resource limitations.}
  \centering   
  \begin{tabular}{l|ccccc}
    \toprule
    \multicolumn{6}{c}{Wiki10-31K} \\
    \midrule
    Method & P@1 & P@2& P@3 & P@5 & R-Prec\\
    \midrule
    HGCLR\textsuperscript{\textdagger} & - & - & - & - & - \\
    HBGL & 53.10 & 52.79 & 53.15 & 52.32 & 26.83 \\
    \midrule
    XR-Transf. & \textbf{88.12} & \textbf{84.87} & \textbf{80.39} & \textbf{70.74} & \textbf{38.93} \\
    CascadeXML & 86.80 & 81.90 & 76.70 & 67.04 & 35.76\\
  \end{tabular}
  \label{tab:wiki10-31k}

  \begin{tabular}{c|ccccc}
    \toprule
    \multicolumn{6}{c}{AmazonCat-13K} \\
    \midrule
    Method & P@1 & P@2 & P@3 & P@5 & R-Prec\\
    \midrule
    HGCLR\textsuperscript{\textdagger} & - & - & - & - & -   \\
    HBGL & 61.01 & 62.21 & 63.10 & 63.20 & 54.67 \\
    \midrule
    XR-Transf. & 96.54 & 90.89 & 83.38 & 67.76 & 82.02 \\
    CascadeXML & \textbf{96.98} & \textbf{91.66} & \textbf{84.54} & \textbf{68.03} & \textbf{83.13}  \\
  \end{tabular}
  \label{tab:amazoncat-13k}

  \begin{tabular}{c|cccccc}
    \toprule
    \multicolumn{6}{c}{Amazon-670K} \\
    \midrule
    Method & P@1 & P@2 & P@3 & P@5 &R-Prec\\
    \midrule
    HGCLR\textsuperscript{\textdagger} & - & - & - & - & - \\
    HBGL\textsuperscript{\textdagger} & - & - & - & - & -\\
    \midrule
    XR-Transf. & \textbf{49.17} & \textbf{46.12} & \textbf{43.75} & \textbf{39.91} & \textbf{39.91} \\
    CascadeXML & 47.86 & 45.10 & 42.88 & 39.15 & 39.66 \\
    \bottomrule
  \end{tabular}
  \label{tab:amazon-670k}
\end{table}

Results on the AmazonCat-13K dataset (see Table~\ref{tab:amazoncat-13k}, middle) show a slightly different picture in the comparison of the XML models. CascadeXML outperforms both XR-Transformer and HBGL, reaching an R-Precision of $83.13$. 
Although the improvement upon XR-Transformer is rather small with $1.11$ points, again, a sizeable gap to HBGL is observed with a difference of $28.46$ points in R-Precision. 
Also, the overall performance for this dataset is high compared to other datasets, reaching values in R-Precision up to $83.13$.

The Amazon-670K dataset (results are shown in Table~\ref{tab:amazon-670k}, bottom) with its extensive label space represents a greater challenge than the previous datasets. 
Unfortunately, we were unable to produce results for HBGL due to resource limitations. 
Even with the adjustments to the segmentation described in Section~\ref{sec:methods:xml_to_htc}, we could not get around the fact that HTC methods are not designed to deal with extremely large label spaces.
Comparing the XML models, XR-Transformer beats CascadeXML across all metrics with an R-Precision of $39.91$, slightly outperforming its counterpart by $0.25$ points.
  
Due to resource limitations, HBGL fails on the Amazon 670k dataset and no results could be produced for HGCLR, too. 
The latter is because HGCLR faces several critical challenges that need to be addressed. The main problem is that the model uses numerous $O(n^2)$ complexity functions in different parts of the code, namely in graph encoding and contrastive learning. 
Here, the inclusion of the extreme number of labels used in XML datasets is likely to cause memory problems on the GPU. 
With the help of the authors of HGCLR, the problem was narrowed down to HGCLR's dense attention functions, which are difficult to optimize for large numbers of labels, which is why experiments with HGCLR on XML datasets did not complete.

\subsection{Ablation: Results of the Leaf-Runs using XML models on HTC datasets}
\label{app:leaf-runs}

The transfer of HTC datasets to the XML world can be achieved in a number of ways. In the HTC world, datasets come with semantic label hierarchies, and if a label at the leaf level of that hierarchy is relevant to a query, this means that all parent labels of that leaf label are implicitly relevant to that query as well. In our main experiments, we included both leaf labels and all implicit labels in the transfer of HTC datasets to the XML domain. In this additional analysis, we explore the other possible way of transferring the dataset, which ignores the implied positive labels derived from the hierarchy and focuses only on those labels that have no further children. 
To enable a fair comparison between XML and HTC models, the datasets were also adjusted for the HTC world, including removing all positive labels that are not leaf labels for each data point. 
It should be noted that this adjusted task is much harder for models from both worlds for two reasons. 
First, the number of relevant labels for each query is lower in this version, which has the consequence that P@k results are naturally worse.
Second, information is lost in the discarding of labels that are not leaf labels in the hierarchy. 

The results of these runs with modified datasets can be seen in Table~\ref{tab:leaf-runs}. 
On the NYT dataset, CascadeXML and HGCLR remain the best-performing models, with HBGL's performance dropping strongly from $75.93$ to $42.30$ points in R-Precision compared to the main analysis. 
On the Web of Science dataset, which is a single-label dataset after the removal of all parent labels, HBGL overtakes HGCLR as the best-performing model with an R-Precision of $80.65$, narrowly beating CascadeXML with an R-Precision of $77.47$. 
XR-Transformer achieves $69.97$ points, and HGCLR is in last place with only $22.74$ points. 
On the RCV1-V2 dataset, HGCLR performs best again, followed by CascadeXML and XR-Transformer. 
HBGL is not able to handle this version of the RCV1 dataset well and achieves an R-Precision of only $8.94$ points. 

\begin{table}[h]
  \centering
  \caption{Results of the Leaf-Runs using the XML models on the HTC datasets}
  \resizebox{0.5\textwidth}{!}{
  \label{tab:leaf-runs}
  \begin{tabular}{c|ccccc}
    \toprule
    \multicolumn{6}{c}{NY-Times} \\
    \midrule
    Method & P@1 & P@2 & P@3 & P@5 & R-Prec\\
    \midrule
    HBGL & 57.46 & 42.73 & 33.10 & 21.97 & 42.30 \\
    HGCLR & 82.93 & 64.74 & 53.51 & 38.82 & 74.09 \\
    \midrule
    XR-Transformer & 82.14 & 62.95 & 51.78 & 37.67 & 72.27 \\
   CascadeXML & 82.89 & 63.75 & 51.99 & 37.90 & 74.32\\
    \toprule
    \multicolumn{6}{c}{Web of Science} \\
    \midrule
    Method & P@1 & P@2 & P@3 & P@5 & R-Prec\\
    \midrule
    HBGL & 80.65 & 40.49 & 26.99 & 16.19 & 80.65 \\
    HGCLR & 22.74 & 20.08 & 18.93 & 19.74 & 22.74\\
    \midrule
    XR-Transformer & 69.97 & 40.53 & 28.53 & 17.86 & 69.97 \\
    CascadeXML & 77.41 & 42.14 & 29.03 & 17.92 & 77.47\\
    \toprule
    \multicolumn{6}{c}{RCV1-V2} \\
    \midrule
    Method & P@1 & P@2 & P@3 & P@5 & R-Prec\\
    \midrule
    HBGL & 13.02 & 7.50 & 5.17 & 3.12 & 8.94 \\
    HGCLR & 92.70 & 59.99 & 43.81 & 27.96 & 83.26 \\
    \midrule
    XR-Transformer & 80.97 & 52.56 & 38.81 & 25.15 & 76.38 \\
    CascadeXML & 86.36 & 55.39 & 40.55 & 26.16 & 81.75\\
    \bottomrule
  \end{tabular}
  }
\end{table}

Overall, the results of HTC models on these adjusted datasets are very inconsistent. This is likely due to their reliance on relational information between labels. Part of this information is lost when only taking leaf labels into account in training and testing. 
XML models do not seem to rely on this information as heavily. 

\section{Discussion}
\label{sec:discussion}
\paragraph{Key Insights}
\label{sec:keyresults}
Our main takeaway is that XML methods are well suited for transfer to the HTC domain and achieve competitive results on HTC datasets.
However, HTC methods are not able to manage the sheer size of labels in the XML datasets (see Section~\ref{sec:results-rq1}). 
XML models and their label-clustering techniques work well on HTC datasets and, due to their inherent method of creating an artificial hierarchy, do not even rely on the taxonomies that come with HTC datasets. On the HTC datasets NYT and RCV1-V2, the result of CascadeXML and HGCLR are very close together in first and second place for the P@1 to P@5 metrics and R-Precision, with CascadeXML slightly better on NYT and HGCLR narrowly winning on RCV1-V2. In other words, a generic (extreme) multi-label classification method is better at hierarchical text classification than methods specifically designed for hierarchical text classification.

The only HTC dataset on which both HTC methods HBGL and HGCLR remain best on the R-Precision metric is the WoS dataset. 
This is likely due to the unique anatomy of this dataset. 
While all the other datasets have variable numbers of relevant labels for each document, see Table~\ref{tab:datasets}, the WoS dataset contains exactly two relevant labels for each data point and also comes with a specific label hierarchy: Every document is assigned a single parent label and a single leaf label. This makes it more similar to single-label datasets than to multi-label datasets, and it may not be suitable for the clustering techniques used in XML models. 

It should also be noted that HGCLR outperforms HBGL on all three datasets from the HTC domain. 
At first glance, this seems to contradict the results presented by the authors of HBGL, which indicate that HBGL is the stronger model of the two, but this can be explained by the different selection of performance metrics in the HBGL paper and this one. 
HBGL produces better F1 scores, and HGCLR is stronger on the P@k metric and R-Precision. This means that HGCLR is better at identifying the top labels but struggles to correctly identify all the relevant labels, resulting in worse F1 scores. 
HBGL seems to be the opposite: Although its top predictions are not as strong as those of HGCLR or CascadeXML, it is better at correctly identifying relevant labels that are less obvious.

For the XML datasets selected from the XML domain, the comparison is not close. 
Neither HBGL nor HGCLR produce competitive results. 
HGCLR is unable to complete the training process on any of the XML datasets due to its $O(n^2)$ complexity, where $n$ is the number of possible labels. 
HBGL faces similar problems and does not produce results on the Amazon-670K dataset, even on a $30k$ subset of training samples. 
For comparison, the largest HTC dataset regarding label space is NYT with $166$ labels. 
Even the smallest XML dataset in terms of labels, AmazonCat-13K, already contains $13,330$ labels. 
While HBGL is able to complete its training and testing pipelines on the smallest datasets from the XML world, AmazonCat-13K and Wiki10-31K, it still performs considerably worse than the models from the XML domain (see Section~\ref{sec:htc-on-xml-datasets}). 
Between these two, XR-Transformer is better on the Wiki10-31K and Amazon-670K datasets, and CascadeXML performs best on AmazonCat-13K. 

\paragraph{Partitioning the Label Space}
Both the HTC and XML methods aim to tackle multi-label text classification and make use of a hierarchy to leverage relational information about the label space. 
However, their approaches to utilizing this information are quite different. 
The hierarchies employed by HTC datasets and methods differ from the hierarchical label trees created prior to the training process of XML models. 
While the former is based on semantic similarities, the latter are characterized by clustering labels together that have similar relevance frequencies and are relevant on documents with similar TF-IDF feature representations. 
The results of our experiments indicate that the former approach, which is oriented towards a human understanding of label similarity, may not be as beneficial for a machine-driven learning process as the latter. 
Furthermore, XML models are particularly strong on the P@1 metric, which could also be due to this difference in clustering strategy. 
Frequency-based clustering encourages the models to predict frequent labels even more often, which is beneficial for predicting head labels and increases P@1, especially when the evaluated dataset contains strong head labels that occur disproportionately, as is the case in the  RCV1-V2 and NYT datasets. 
This is also supported by the results of HBGL** where the native taxonomy was replaced by an XML-style one. 
Although the improvements are too marginal to claim that this version of clustering is generally superior, they are best in P@1.

\paragraph{Metrics}

As mentioned in Section~\ref{sec:measures}, the XML and HTC domains use different measures to evaluate model performance. 
In the domain of HTC, F1 scores are mainly used, and in the domain of XML, the P@k metric is predominant. However, the results of our experiments suggest that it is not sufficient to use only one or the other to capture all aspects of model performance.
Models should ideally be evaluated on all of these metrics and possibly even the R-Precision metric. 

As presented by the authors of HBGL, HBGL outperforms HGCLR on both the Micro-F1 and Macro-F1 metrics, but in our experiments, HGCLR achieved better results on the P@k and R-Precision metrics, \textbf{suggesting that HBGL is not stronger than HGCLR per se, but that the two models are rather good at different aspects of text classification}. 
A comparison based on just one set of these metrics would have missed this observation. 
It would, therefore, be beneficial to make comparisons using a wider range of metrics.
Note that some models would need to be adapted to support both families of performance measures, as F1-scores are based on yes or no classifications, and both P@k and R-Precision scores are based on rankings. The system under test would have to be able to support both paradigms (see Section~\ref{sec:measures}).

\paragraph{Threats to Validity}
There are a number of threats to validity that need to be discussed. The first relates to the structure of some of the datasets used. 
The dataset transfer between XML and HTC methods may not be optimal due to differences in their hierarchical structures. XML datasets do not inherently contain hierarchical labels. 
Instead, they are generated through clustering. These clusters include meta labels, which guide the learning process towards the actual labels used for classification, the leaf nodes.
If we apply HTC methods on such generated hierarchies, as in Appendices~\ref{sec:methods:xml_to_htc} and \ref{sec:results-rq1}, the HTC methods expect the entire hierarchy to consist of labels valid for classification. 
Since only the leaf nodes are actual labels and the rest are meta labels, HTC models are limited to classifying only at the final level of the hierarchy. In other words, the XML datasets are actually only multi-label classification tasks, where the generated hierarchical label tree is a tool and not used to classify the documents. 

The performance of CascadeXML reported by~\citet{kharbanda2022cascadexml} on Wiki10-31k and Amazon-670k could not be achieved. 
On the Wiki10-31k dataset, our models exhibit a mean deficit of $3.18$ points across the P@1, P@3, and P@5 metrics. 
On the Amazon-670k dataset, it is $3.76$ points. 
This might be due to CascadeXML being run in an ensemble setting while we employ the BERT-only variant. 
However, on AmazonCat-13K our numbers of CascadeXML are actually better than the ensemble.
The outcome of our analysis comparing HTC versus XML methods on the respective datasets is not affected because CascadeXML with BERT outperforms XR-Transformer on all HTC datasets and the AmazonCat-13k dataset.

Detailed hyperparameter optimization, \eg in the form of a grid search, was out of scope due to the number of experiments. Instead, we used the hyperparameters tuned by the authors of the respective methods. 
For the transfer from HTC to XML, we used the hyperparameters of the datasets that came closest to those of the other domain. 
These were mostly those from Wiki10-31k in the case of XML datasets for HTC methods and those from RCV1-V2 (HBGL configuration) in the case of HTC datasets for XML methods.

In summary, our results show that XML methods, in particular CascadeXML, are very competitive and can be considered for hierarchical text classification tasks. 
CascadeXML works particularly well on the NYT dataset, even outperforming state-of-the-art HTC methods such as HGCLR and HBGL. 
This suggests that the structure of the NYT dataset, which involves complex and deep hierarchical categorizations of news articles, is particularly well suited to the strengths of XML methods. 
At the same time, despite the challenging comparison, HGCLR has shown that HTC methods remain relevant and effective in certain scenarios. 
The results suggest that HTC methods could benefit from adaptation to larger datasets, which would require refinements to cope with increasing complexity.

\section{Conclusion}
\label{sec:conclusion}

Our experiments show that XML methods are very competitive and should be considered for HTC tasks. 
In particular, CascadeXML performed well and, on the NYT dataset, outperformed the state-of-the-art HTC methods HGCLR and HBGL. 
The XML methods have the conceptual advantage that they do not require a pre-defined label hierarchy but instead construct an artificial hierarchy to partition the label space, which makes them applicable to a wider range of multi-label classification problems, including hierarchical text classification as shown here.
In the other direction, we found that HTC models are hardly applicable to the XML datasets due to limited scalability.
Our findings suggest that future work on HTC should compare to XML methods and that ranking-based metrics P@k and R-Precision should be considered for a more thorough comparison of methods. 

\arxivonly{
We investigated the cross-domain capabilities of Hierarchical Text Classification (HTC) and eXtreme Multi-Label Classification (XML) models, using cross-dataset transfer to evaluate their performance on unfamiliar datasets and label spaces. 
Our cross-domain comparison shows that XML models, in particular CascadeXML, demonstrate considerable adaptability and competitive accuracy when applied to HTC datasets, effectively handling smaller hierarchical structures despite being designed for large, unstructured label pools. 
}

\paragraph{Limitations}
We do not evaluate the training time of the various models, as the original papers lack sufficient information, and hardware configurations were adjusted during experimentation to suit specific model-dataset combinations. 
This was particularly necessary for running the HTC models on the XML datasets, in the case of HGCLR, with the help of the authors, but unsuccessful. 
Nevertheless, we have noted that XML models train much faster than HTC models.

\arxivonly{
Like the existing models in HTC and XML discussed in Section~\ref{sec:relatedwork}, the datasets used in this study are exclusively in the English language. 
Consequently, the results presented in this work are solely applicable to the English language domain and may not be directly transferable to other languages.
}

\arxivonly{
\section{Ethical Considerations}

While text classification has a broad spectrum of application domains ranging from recommender systems and information retrieval to tasks such as sentiment analysis, e-commerce, advertising, and news classification~\cite{textclassificationsurvey,intro3,intro4,htc}, we believe that no specific societal consequences require immediate emphasis in the context of this work comparing HTC with XML models.
}

\begin{ack}
The authors acknowledge support from the state of Baden-Württemberg through bwHPC.
\end{ack}

\bibliography{references}

@article{DBLP:journals/kbs/ZhangLSXTH24-halb,
  author       = {Jun Zhang and
                  Yubin Li and
                  Fanfan Shen and
                  Chenxi Xia and
                  Hai Tan and
                  Yanxiang He},
  title        = {Hierarchy-Aware and Label Balanced Model for Hierarchical Text Classification},
  journal      = {Knowl. Based Syst.},
  volume       = {300},
  pages        = {112153},
  year         = {2024},
 no_url = {https://doi.org/10.1016/j.knosys.2024.112153},
  doi          = {10.1016/j.knosys.2024.112153},
  bibsource    = {dblp computer science bibliography, https://dblp.org}
}

@inproceedings{hill2024,
  author       = {He Zhu and
                  Junran Wu and
                  Ruomei Liu and
                  Yue Hou and
                  Ze Yuan and
                  Shangzhe Li and
                  Yicheng Pan and
                  Ke Xu},
  no_editor       = {Kevin Duh and
                  Helena G{\'{o}}mez{-}Adorno and
                  Steven Bethard},
  title        = {{HILL:} {H}ierarchy-aware Information Lossless Contrastive Learning for Hierarchical Text Classification},
  booktitle    = {NAACL 2024},
  pages        = {4731--4745},
  publisher    = {ACL},
  year         = {2024},
 no_url = {https://doi.org/10.18653/v1/2024.naacl-long.265},
  doi          = {10.18653/v1/2024.naacl-long.265},
  bibsource    = {dblp computer science bibliography, https://dblp.org}
}

@article{DBLP:journals/tsp/SchusterP97,
  author       = {Mike Schuster and
                  Kuldip K. Paliwal},
  title        = {Bidirectional recurrent neural networks},
  journal      = {{IEEE} Trans. Signal Process.},
  volume       = {45},
  number       = {11},
  pages        = {2673--2681},
  year         = {1997},
 no_url = {https://doi.org/10.1109/78.650093},
  doi          = {10.1109/78.650093},
  bibsource    = {dblp computer science bibliography, https://dblp.org}
}

@article{DBLP:journals/corr/abs-2307-09288-llama2,
  author       = {Hugo Touvron and
                  Louis Martin and
                  Kevin Stone and
                  Peter Albert and
                  Amjad Almahairi and
                  Yasmine Babaei and
                  Nikolay Bashlykov and
                  Soumya Batra and
                  Prajjwal Bhargava and
                  Shruti Bhosaleand
                 others},
  title        = {Llama 2: Open Foundation and Fine-Tuned Chat Models},
  journal      = {CoRR},
  volume       = {abs/2307.09288},
  year         = {2023},
 no_url = {https://doi.org/10.48550/arXiv.2307.09288},
  no_doi          = {10.48550/ARXIV.2307.09288},
  eprinttype    = {arXiv},
  eprint       = {2307.09288},
  bibsource    = {dblp computer science bibliography, https://dblp.org}
}

@article{DBLP:journals/corr/abs-2303-08774-gpt4,
  author       = {OpenAI},
  title        = {{GPT-4} Technical Report},
  journal      = {CoRR},
  volume       = {abs/2303.08774},
  year         = {2023},
 no_url = {https://doi.org/10.48550/arXiv.2303.08774},
  no_doi          = {10.48550/ARXIV.2303.08774},
  eprinttype    = {arXiv},
  eprint       = {2303.08774},
  bibsource    = {dblp computer science bibliography, https://dblp.org}
}

@inproceedings{hgclr,
    title = "Incorporating Hierarchy into Text Encoder: a Contrastive Learning Approach for Hierarchical Text Classification",
    author = "Wang, Zihan  and
      Wang, Peiyi  and
      Huang, Lianzhe  and
      Sun, Xin  and
      Wang, Houfeng",
    no_editor = "Muresan, Smaranda  and
      Nakov, Preslav  and
      Villavicencio, Aline",
    booktitle = "ACL",
    no_month = may,
    year = "2022",
    no_address = "Dublin, Ireland",
    no_publisher = "Association for Computational Linguistics",
   no_url = "https://aclanthology.org/2022.acl-long.491",
    doi = "10.18653/v1/2022.acl-long.491",
    no_pages = "7109--7119",
}

@inproceedings{parabe2018,
author = {Prabhu, Yashoteja and Kag, Anil and Harsola, Shrutendra and Agrawal, Rahul and Varma, Manik},
title = {Parabel: Partitioned Label Trees for Extreme Classification with Application to Dynamic Search Advertising},
year = {2018},
no_isbn = {9781450356398},
no_publisher = {International World Wide Web Conferences Steering Committee},
no_address = {Republic and Canton of Geneva, CHE},no_url = {https://doi.org/10.1145/3178876.3185998},
no_doi = {10.1145/3178876.3185998},
booktitle = {WWW},
pages = {993–1002},
numpages = {10},
location = {Lyon, France},
no_series = {WWW '18}
}

@inproceedings{sparselocalembeddings2015,
 author = {Bhatia, Kush and Jain, Himanshu and Kar, Purushottam and Varma, Manik and Jain, Prateek},
 booktitle = {NeurIPS},
 no_editor = {C. Cortes and N. Lawrence and D. Lee and M. Sugiyama and R. Garnett},
 pages = {},
 no_publisher = {Curran Associates, Inc.},
 title = {Sparse Local Embeddings for Extreme Multi-label Classification},
no_url = {https://proceedings.neurips.cc/paper_files/paper/2015/file/35051070e572e47d2c26c241ab88307f-Paper.pdf},
 volume = {28},
 year = {2015}
}

@article{graphormer,
  author       = {Chengxuan Ying and
                  Tianle Cai and
                  Shengjie Luo and
                  Shuxin Zheng and
                  Guolin Ke and
                  Di He and
                  Yanming Shen and
                  Tie{-}Yan Liu},
  title        = {Do Transformers Really Perform Bad for Graph Representation?},
  journal      = {CoRR},
  volume       = {abs/2106.05234},
  year         = {2021},
 no_url = {https://arxiv.org/abs/2106.05234},
  eprinttype    = {arXiv},
  eprint       = {2106.05234},
  bibsource    = {dblp computer science bibliography, https://dblp.org}
}

@inproceedings{hbgl,
    title = "Exploiting Global and Local Hierarchies for Hierarchical Text Classification",
    author = "Jiang, Ting  and
      Wang, Deqing  and
      Sun, Leilei  and
      Chen, Zhongzhi  and
      Zhuang, Fuzhen  and
      Yang, Qinghong",
    no_editor = "Goldberg, Yoav  and
      Kozareva, Zornitsa  and
      Zhang, Yue",
    booktitle = "EMNLP",
    no_month = dec,
    year = "2022",
    no_address = "Abu Dhabi, United Arab Emirates",
    no_publisher = "Association for Computational Linguistics",
   no_url = "https://aclanthology.org/2022.emnlp-main.268",
    doi = "10.18653/v1/2022.emnlp-main.268",
    no_pages = "4030--4039",
}

@article{s2s-ft,
  author       = {Hangbo Bao and
                  Li Dong and
                  Wenhui Wang and
                  Nan Yang and
                  Furu Wei},
  title        = {s2s-ft: Fine-Tuning Pretrained Transformer Encoders for Sequence-to-Sequence
                  Learning},
  journal      = {CoRR},
  volume       = {abs/2110.13640},
  year         = {2021},
 no_url = {https://arxiv.org/abs/2110.13640},
  eprinttype    = {arXiv},
  eprint       = {2110.13640},
  bibsource    = {dblp computer science bibliography, https://dblp.org}
}

@inproceedings{hiagm,
  author       = {Jie Zhou and
                  Chunping Ma and
                  Dingkun Long and
                  Guangwei Xu and
                  Ning Ding and
                  Haoyu Zhang and
                  Pengjun Xie and
                  Gongshen Liu},
  no_editor       = {Dan Jurafsky and
                  Joyce Chai and
                  Natalie Schluter and
                  Joel R. Tetreault},
  title        = {Hierarchy-Aware Global Model for Hierarchical Text Classification},
  booktitle    = {ACL},
  no_pages        = {1106--1117},
  no_publisher    = {Association for Computational Linguistics},
  year         = {2020},
 no_url = {https://doi.org/10.18653/v1/2020.acl-main.104},
  no_doi          = {10.18653/v1/2020.acl-main.104},
  bibsource    = {dblp computer science bibliography, https://dblp.org}
}

@inproceedings{hpt,
  author       = {Zihan Wang and
                  Peiyi Wang and
                  Tianyu Liu and
                  Binghuai Lin and
                  Yunbo Cao and
                  Zhifang Sui and
                  Houfeng Wang},
  no_editor       = {Yoav Goldberg and
                  Zornitsa Kozareva and
                  Yue Zhang},
  title        = {{HPT:} Hierarchy-aware Prompt Tuning for Hierarchical Text Classification},
  booktitle    = {EMNLP},
  no_pages        = {3740--3751},
  no_publisher    = {Association for Computational Linguistics},
  year         = {2022},
 no_url = {https://doi.org/10.18653/v1/2022.emnlp-main.246},
  doi          = {10.18653/V1/2022.EMNLP-MAIN.246},
  bibsource    = {dblp computer science bibliography, https://dblp.org}
}

@inproceedings{gat,
  author       = {Petar Velickovic and
                  Guillem Cucurull and
                  Arantxa Casanova and
                  Adriana Romero and
                  Pietro Li{\`{o}} and
                  Yoshua Bengio},
  title        = {Graph Attention Networks},
  booktitle    = {ICLR},
  no_publisher    = {OpenReview.net},
  year         = {2018},
 no_url = {https://openreview.net/forum?id=rJXMpikCZ},
  bibsource    = {dblp computer science bibliography, https://dblp.org}
}

@article{htc_review,
  title={Hierarchical Text Classification: a review of current research},
  author={Zangari, Alessandro and Marcuzzo, Matteo and Schiavinato, Michele and Rizzo, Matteo and Gasparetto, Andrea and Albarelli, Andrea and others},
  journal={Expert Systems with Applications},
  volume={224},
  year={2023}
}

@inproceedings{yang2020xlnet,
author = {Yang, Zhilin and Dai, Zihang and Yang, Yiming and Carbonell, Jaime and Salakhutdinov, Ruslan and Le, Quoc V.},
title = {XLNet: generalized autoregressive pretraining for language understanding},
year = {2019},
no_publisher = {Curran Associates Inc.},
no_address = {Red Hook, NY, USA},
booktitle = {NeurIPS},
no_articleno = {517},
no_numpages = {11}
}

@article{liu2019roberta,
  title={RoBERTa: A Robustly Optimized BERT Pretraining Approach},
  author={Yinhan Liu and Myle Ott and Naman Goyal and Jingfei Du and Mandar Joshi and Danqi Chen and Omer Levy and Mike Lewis and Luke Zettlemoyer and Veselin Stoyanov},
  journal={ArXiv},
  year={2019},
  volume={abs/1907.11692},
 no_url ={https://api.semanticscholar.org/CorpusID:198953378}
}

@inproceedings{kharbanda2022cascadexml,
  title={{CascadeXML}: Rethinking Transformers for End-to-end Multi-resolution Training in Extreme Multi-label Classification},
  author={Kharbanda, Siddhant and Banerjee, Atmadeep and Schultheis, Erik and Babbar, Rohit},
  booktitle={NeurIPS},
  no_volume={35},
  no_pages={2074--2087},
  year={2022}
}

@inproceedings{xr-transformer,
author = {Zhang, Jiong and Chang, Wei-cheng and Yu, Hsiang-fu and Dhillon, Inderjit S.},
title = {Fast multi-resolution transformer fine-tuning for extreme multi-label text classification},
year = {2024},
no_isbn = {9781713845393},
no_publisher = {Curran Associates Inc.},
no_address = {Red Hook, NY, USA},
booktitle = {NeurIPS},
no_articleno = {556},
no_numpages = {14},
no_series = {NIPS '21}
}

@inproceedings{devlin2019bert,
  author       = {Jacob Devlin and
                  Ming{-}Wei Chang and
                  Kenton Lee and
                  Kristina Toutanova},
  no_editor       = {Jill Burstein and
                  Christy Doran and
                  Thamar Solorio},
  title        = {{BERT:} {P}re-training of Deep Bidirectional Transformers for Language
                  Understanding},
  booktitle    = {NAACL-HLT},
  no_pages        = {4171--4186},
  no_publisher    = {Association for Computational Linguistics},
  year         = {2019},
 no_url = {https://doi.org/10.18653/v1/n19-1423},
  doi          = {10.18653/V1/N19-1423},
  bibsource    = {dblp computer science bibliography, https://dblp.org}
}

@inproceedings{longtaildistribution,
  author       = {Xinli Yue and
                  Ningping Mou and
                  Qian Wang and
                  Lingchen Zhao},
  title        = {Revisiting Adversarial Training Under Long-Tailed Distributions},
  booktitle    = {CVPR},
  no_pages        = {24492--24501},
  no_publisher    = {{IEEE}},
  year         = {2024},
 no_url = {https://doi.org/10.1109/CVPR52733.2024.02312},
  doi          = {10.1109/CVPR52733.2024.02312},
  bibsource    = {dblp computer science bibliography, https://dblp.org}
}

@inproceedings{f-score,
  title={A probabilistic interpretation of precision, recall and F-score, with implication for evaluation},
  author={Goutte, Cyril and Gaussier, Eric},
  booktitle={ECIR},
  no_pages={345--359},
  year={2005},
  no_organization={Springer}
}

@inproceedings{babbar2016dismec,
  author       = {Rohit Babbar and
                  Bernhard Sch{\"{o}}lkopf},
  no_editor       = {Maarten de Rijke and
                  Milad Shokouhi and
                  Andrew Tomkins and
                  Min Zhang},
  title        = {{DiSMEC}: Distributed Sparse Machines for Extreme Multi-label Classification},
  booktitle    = {WSDM},
  pages        = {721--729},
  publisher    = {{ACM}},
  year         = {2017},
 no_url = {https://doi.org/10.1145/3018661.3018741},
  doi          = {10.1145/3018661.3018741},
  bibsource    = {dblp computer science bibliography, https://dblp.org}
}

@inproceedings{ppdsparse,
  author       = {Ian En{-}Hsu Yen and
                  Xiangru Huang and
                  Wei Dai and
                  Pradeep Ravikumar and
                  Inderjit S. Dhillon and
                  Eric P. Xing},
  title        = {{PPDsparse}: {A} Parallel Primal-Dual Sparse Method for Extreme Classification},
  booktitle    = {KDD},
  no_pages        = {545--553},
  no_publisher    = {{ACM}},
  year         = {2017},
 no_url = {https://doi.org/10.1145/3097983.3098083},
  doi          = {10.1145/3097983.3098083},
  bibsource    = {dblp computer science bibliography, https://dblp.org}
}

@inproceedings{you2019attentionxml,
  author       = {Ronghui You and
                  Zihan Zhang and
                  Ziye Wang and
                  Suyang Dai and
                  Hiroshi Mamitsuka and
                  Shanfeng Zhu},
  no_editor       = {Hanna M. Wallach and
                  Hugo Larochelle and
                  Alina Beygelzimer and
                  Florence d'Alch{\'{e}}{-}Buc and
                  Emily B. Fox and
                  Roman Garnett},
  title        = {AttentionXML: Label Tree-based Attention-Aware Deep Model for High-Performance
                  Extreme Multi-Label Text Classification},
  booktitle    = {NeurIPS},
  no_pages        = {5812--5822},
  year         = {2019},
 no_url = {https://proceedings.neurips.cc/paper/2019/hash/9e6a921fbc428b5638b3986e365d4f21-Abstract.html},
  bibsource    = {dblp computer science bibliography, https://dblp.org}
}

@inproceedings{kharbanda2022inceptionxml,
  author       = {Siddhant Kharbanda and
                  Atmadeep Banerjee and
                  Devaansh Gupta and
                  Akash Palrecha and
                  Rohit Babbar},
  no_editor       = {Hsin{-}Hsi Chen and
                  Wei{-}Jou (Edward) Duh and
                  Hen{-}Hsen Huang and
                  Makoto P. Kato and
                  Josiane Mothe and
                  Barbara Poblete},
  title        = {{InceptionXML}: {A} Lightweight Framework with Synchronized Negative
                  Sampling for Short Text Extreme Classification},
  booktitle    = {SIGIR},
  no_pages        = {760--769},
  no_publisher    = {{ACM}},
  year         = {2023},
 no_url = {https://doi.org/10.1145/3539618.3591699},
  doi          = {10.1145/3539618.3591699},
  bibsource    = {dblp computer science bibliography, https://dblp.org}
}

@article{galke2023really,
  title={Are We Really Making Much Progress in Text Classification? {A} Comparative Review},
  author={Galke, Lukas and Diera, Andor and Lin, Bao Xin and Khera, Bhakti and Meuser, Tim and Singhal, Tushar and Karl, Fabian and Scherp, Ansgar},
  journal={arXiv preprint arXiv:2204.03954},
  year={2022}
}

@article{sun2023text,
  author       = {Xiaofei Sun and
                  Xiaoya Li and
                  Jiwei Li and
                  Fei Wu and
                  Shangwei Guo and
                  Tianwei Zhang and
                  Guoyin Wang},
  title        = {Text Classification via Large Language Models},
  journal      = {CoRR},
  volume       = {abs/2305.08377},
  year         = {2023},
 no_url = {https://doi.org/10.48550/arXiv.2305.08377},
  no_doi          = {10.48550/ARXIV.2305.08377},
  eprinttype    = {arXiv},
  eprint       = {2305.08377},
  bibsource    = {dblp computer science bibliography, https://dblp.org}
}

@article{lightxml,
  author       = {Ting Jiang and
                  Deqing Wang and
                  Leilei Sun and
                  Huayi Yang and
                  Zhengyang Zhao and
                  Fuzhen Zhuang},
  title        = {LightXML: Transformer with Dynamic Negative Sampling for High-Performance
                  Extreme Multi-label Text Classification},
  journal      = {CoRR},
  volume       = {abs/2101.03305},
  year         = {2021},
 no_url = {https://arxiv.org/abs/2101.03305},
  eprinttype    = {arXiv},
  eprint       = {2101.03305},
  bibsource    = {dblp computer science bibliography, https://dblp.org}
}

@Article{textclassificationsurvey,
AUTHOR = {Gasparetto, Andrea and Marcuzzo, Matteo and Zangari, Alessandro and Albarelli, Andrea},
TITLE = {A Survey on Text Classification Algorithms: From Text to Predictions},
JOURNAL = {Information},
VOLUME = {13},
YEAR = {2022},
NUMBER = {2},
ARTICLE-NUMBER = {83},no_url = {https://www.mdpi.com/2078-2489/13/2/83},
no_ISSN = {2078-2489},
DOI = {10.3390/info13020083}
}

@inproceedings{htc,
  author={Aixin Sun and Ee-Peng Lim},
  booktitle={ICDM}, 
  title={Hierarchical text classification and evaluation}, 
  year={2001},
  volume={},
  number={},
  pages={521-528},
  doi={10.1109/ICDM.2001.989560}

}

@inproceedings{intro3,
  title={Extreme multi-label learning for semantic matching in product search},
  author={Chang, Wei-Cheng and Jiang, Daniel and Yu, Hsiang-Fu and Teo, Choon Hui and Zhang, Jiong and Zhong, Kai and Kolluri, Kedarnath and Hu, Qie and Shandilya, Nikhil and Ievgrafov, Vyacheslav and others},
  booktitle={Proceedings of the 27th ACM SIGKDD Conference on Knowledge Discovery \& Data Mining},
  pages={2643--2651},
  year={2021}
}

@inproceedings{intro4,
  title={Slice: {S}calable linear extreme classifiers trained on 100 million labels for related searches},
  author={Jain, Himanshu and Balasubramanian, Venkatesh and Chunduri, Bhanu and Varma, Manik},
  booktitle={WSDM},
  no_pages={528--536},
  year={2019}
}

@article{li2021survey,
  author       = {Qian Li and
                  Hao Peng and
                  Jianxin Li and
                  Congying Xia and
                  Renyu Yang and
                  Lichao Sun and
                  Philip S. Yu and
                  Lifang He},
  title        = {A Survey on Text Classification: From Shallow to Deep Learning},
  journal      = {CoRR},
  volume       = {abs/2008.00364},
  year         = {2020},
 no_url = {https://arxiv.org/abs/2008.00364},
  eprinttype    = {arXiv},
  eprint       = {2008.00364},
  bibsource    = {dblp computer science bibliography, https://dblp.org}
}

@article{yu2023open,
  author       = {Hao Yu and
                  Zachary Yang and
                  Kellin Pelrine and
                  Jean{-}Fran{\c{c}}ois Godbout and
                  Reihaneh Rabbany},
  title        = {Open, Closed, or Small Language Models for Text Classification?},
  journal      = {CoRR},
  volume       = {abs/2308.10092},
  year         = {2023},
 no_url = {https://doi.org/10.48550/arXiv.2308.10092},
  no_doi          = {10.48550/ARXIV.2308.10092},
  eprinttype    = {arXiv},
  eprint       = {2308.10092},
  bibsource    = {dblp computer science bibliography, https://dblp.org}
}

@misc{ds-wiki1031,
      title={Enhancing Navigation on Wikipedia with Social Tags}, 
      author={Arkaitz Zubiaga},
      year={2012},
      eprint={1202.5469},
      archivePrefix={arXiv},
      primaryClass={cs.IR}
}

@inproceedings{ds-amazon,
author = {McAuley, Julian and Leskovec, Jure},
title = {Hidden factors and hidden topics: understanding rating dimensions with review text},
year = {2013},
no_isbn = {9781450324090},
no_publisher = {Association for Computing Machinery},
no_address = {New York, NY, USA},no_url = {https://doi.org/10.1145/2507157.2507163},
doi = {10.1145/2507157.2507163},
booktitle = {RecSys},
no_pages = {165–172},
no_numpages = {8},
no_location = {Hong Kong, China},
no_series = {RecSys '13}
}

@inproceedings{ds-wos,
   title={HDLTex: Hierarchical Deep Learning for Text Classification},
  no_url ={http://dx.doi.org/10.1109/ICMLA.2017.0-134},
   DOI={10.1109/icmla.2017.0-134},
   booktitle={2017 16th IEEE International Conference on Machine Learning and Applications (ICMLA)},
   publisher={IEEE},
   author={Kowsari, Kamran and Brown, Donald E. and Heidarysafa, Mojtaba and Jafari Meimandi, Kiana and Gerber, Matthew S. and Barnes, Laura E.},
   year={2017},
   month=dec }

@inproceedings{ds-nyt,
    title = "{HFT}-{CNN}: Learning Hierarchical Category Structure for Multi-label Short Text Categorization",
    author = "Shimura, Kazuya  and
      Li, Jiyi  and
      Fukumoto, Fumiyo",
    no_editor = "Riloff, Ellen  and
      Chiang, David  and
      Hockenmaier, Julia  and
      Tsujii, Jun{'}ichi",
    booktitle = "EMNLP",
    year = "2018",
    no_address = "Brussels, Belgium",
    no_publisher = "Association for Computational Linguistics",
   no_url = "https://aclanthology.org/D18-1093",
    doi = "10.18653/v1/D18-1093",
    no_pages = "811--816",
}

@article{ds-rcv1v2,
author = {Lewis, David and Yang, Yiming and Russell-Rose, Tony and Li, Fan},
year = {2004},
month = {04},
pages = {361-397},
title = {RCV1: A New Benchmark Collection for Text Categorization Research.},
volume = {5},
journal = {Journal of Machine Learning Research}
}

@inproceedings{eval-in-ir,
    title = "Evaluation in information retrieval",
    author = "Manning, Christopher and Raghavan, Prabhakar and Schütz, Hinrich",
    editor = "Jurafsky, Dan  and
      Chai, Joyce  and
      Schluter, Natalie  and
      Tetreault, Joel",
    booktitle = "Introuctino to Information Retrieval",
    year = "2009",
    address = "Online",
    publisher = "University of Stamford Press",
   no_url = "https://nlp.stanford.edu/IR-book/pdf/08eval.pdf",
    pages = "151-175",
    
}

@inproceedings{htc-info-max,
    title = {{HTCInfoMax}: A Global Model for Hierarchical Text Classification via Information Maximization},
    author = "Deng, Zhongfen and 
              Peng, Hao and
              He, Dongxiao and 
              Li, Jianxin and 
              Yu, Philip",
    booktitle = "NAACL-HLT",
    no_month = jun, 
    no_year = "2021",
    no_address = "Online", 
    no_publisher = "Association for Computational Linguistics", 
   no_url = "https://aclanthology.org/2021.naacl-main.260", 
    doi = "10.18653/v1/2021.naacl-main.260", 
    no_pages = "3259–3265"
}

@inproceedings{himatch,
    title = {Hierarchy-aware Label Semantics Matching Network for Hierarchical Text Classification},
    author = "Chen, Haibin and 
                Ma, Qianli and 
                Lin, Zhenxi and 
                Yan, Jiangyue",
    booktitle ="ACL/IJCNLP" ,
    no_month = aug,
    year = "2021",
    no_address = "Online",
    no_publisher = "Association for Computational Linguistics",
   no_url = "https://aclanthology.org/2021.acl-long.337",
    doi = "10.18653/v1/2021.acl-long.337",
    pages = "4370–4379"
}

@inproceedings{XMCsparse,
  title={A no-regret generalization of hierarchical softmax to extreme multi-label classification},
  author={Wydmuch, Marek and Jasinska, Kalina and Kuznetsov, Mikhail and Busa-Fekete, R{\'o}bert and Dembczynski, Krzysztof},
  booktitle={NeurIPS},
  no_volume={31},
  year={2018}
}

@inproceedings{PSPatK,
author = {Jain, Himanshu and Prabhu, Yashoteja and Varma, Manik},
title = {Extreme Multi-label Loss Functions for Recommendation, Tagging, Ranking \& Other Missing Label Applications},
year = {2016},
no_isbn = {9781450342322},
no_publisher = {Association for Computing Machinery},
no_address = {New York, NY, USA},no_url = {https://doi.org/10.1145/2939672.2939756},
doi = {10.1145/2939672.2939756},
booktitle = {KDD},
no_pages = {935–944},
no_numpages = {10},
no_location = {San Francisco, California, USA},
no_series = {KDD '16}
}

@inproceedings{tang2009large,
  title={Large scale multi-label classification via metalabeler},
  author={Tang, Lei and Rajan, Suju and Narayanan, Vijay K},
  booktitle={WWW},
  pages={211--220},
  year={2009}
}

\arxivonly{
\clearpage
\appendix

\section*{Supplementary Materials}

\section{Further Discussions}

In some of the XML datasets, texts are labeled with arbitrary labels that have no significant semantic meaning (\ie "!", "!!" , "AE04752"). 
While the lack of semantic context in these labels may raise concerns about the overall suitability of such datasets for hierarchical classification, it is unlikely a problem for our purposes. 
Some of the evaluated models, like HBGL, can perform better by incorporating label semantics (using pre-trained embeddings of the textual representation of the label). 
However, ablation studies for HBGL~\cite{hbgl} and HGCLR~\cite{hgclr} show that the impact of label semantics is relatively small.

Some datasets commonly used in the XML world, such as Wiki-500K and Amazon-3M, were included in the experimental setting in the first place but could not be tested for XR-Transformer and HTC methods due to resource limitations. 
We ran our experiments on NVIDIA H100 80GB GPUs with 1TB of RAM available but encountered OOM errors when training XR-Transformer on Wiki-500K and Amazon-3M. 
For the HTC methods, training time was the limiting factor. We were unable to complete the training within the 48-hour window imposed by our computing resource provider.
}

\clearpage

\begin{table}[t]
\small
\centering
\caption{Results for the HTC datasets, namely New York Times Annotated Corpus (top), Web of Science (middle), Reuters Corpus Volume 1 (bottom).
HBGL** indicates that the semantic hierarchy, as provided by the datasets in HTC, has been replaced by a synthetic hierarchical label tree as it is computed in XML.}
\label{tab:htc_results}

\resizebox{0.5\textwidth}{!}{
\begin{tabular}{l|cccc|cc}
  \toprule
  \multicolumn{7}{c}{NY-Times} \\
  \midrule
  Method & R-Prec & P@1 & P@3 & P@5 & F1-Micro & F1-Macro \\
  \midrule
  HGCLR & 84.22 & 95.02  &  83.98 &  71.74 & 78.86 & 67.96 \\
  HBGL & 75.93 & 83.81 & 75.37 & 64.87& 80.01 & \textbf{70.14}\\
  RADAr** & 79.79& 94.85 & 81.36 & 68.61 & 79.30 & 67.68 \\
  \midrule
  XR-Transf. & \textbf{86.04} & \textbf{95.78} & \textbf{85.37} & \textbf{72.91} & 78.45 & 63.13\\
  CascadeXML & 85.60 & 95.35 & 84.79 & 72.54 & \textbf{80.35} & 69.79 \\
  \bottomrule
\end{tabular}
}

\vspace{1em}

\resizebox{0.5\textwidth}{!}{
\begin{tabular}{l|cccc|cc}
  \toprule
  \multicolumn{7}{c}{Web of Science} \\
  \midrule
  Method & R-Prec & P@1 & P@3 & P@5 & F1-Micro & F1-Macro \\
  \midrule
  HGCLR & 86.44 & 91.05 & \textbf{60.27} & \textbf{37.48} & 87.11 & 81.20 \\
  HBGL & 86.02 & 85.97 & 57.62 & 34.57& \textbf{87.68} & \textbf{82.01} \\
  RADAr** & 86.65& \textbf{91.47}& 57.77 & 34.66 & 86.65 & 81.10 \\
  \midrule
  XR-Transf. & 80.56 & 88.52 & 57.89 & 36.43 & 77.74 & 61.73\\
  CascadeXML & \textbf{86.91} & 91.27 & 59.99 & 37.06 & 87.26 & 81.64 \\
  \bottomrule
\end{tabular}
}

\vspace{1em}

\resizebox{0.5\textwidth}{!}{
\begin{tabular}{l|cccc|cc}
  \toprule
  \multicolumn{7}{c}{RCV1-V2} \\
  \midrule
  Method & R-Prec & P@1 & P@3 & P@5 & F1-Micro & F1-Macro \\
  \midrule
  HGCLR & 90.49 & 96.78  &  83.40 &  57.86  & 86.49 & 68.31 \\
  HBGL & 85.19 & 91.51 & 80.43 & 54.57 & 86.94 & \textbf{70.49} \\
  RADAr** & 86.80 & 96.61 & 81.46 & 54.70 & \textbf{87.26} & 69.07 \\
  \midrule
  XR-Transf. & 90.36 & \textbf{97.40} & \textbf{83.66} & 57.47 & 85.97 & 60.44 \\
  CascadeXML & \textbf{90.81} & 96.64 & 83.63 & \textbf{57.99} & 87.08 & 69.14 \\
  \bottomrule
\end{tabular}
}

\end{table}

\end{document}